# A Tangent Distance Preserving Dimensionality Reduction Algorithm

Xu Zhao
The Laboratory of Computer Software and Theory
Beijing University of Technology
PingLeYuan 100, ChaoYang District, Beijing,
PR China
zhaoxu166@gmail.com

Zongli Jiang
The Laboratory of Computer Software and Theory
Beijing University of Technology
PingLeYuan 100, ChaoYang District, Beijing,
PR China
jiangzl@bjut.edu.cn

**ABSTRACT**
This paper considers the problem of nonlinear dimensionality reduction. Unlike existing methods, such as LLE, ISOMAP, which attempt to unfold the true manifold in the low dimensional space, our algorithm tries to preserve the nonlinear structure of the manifold, and shows how the manifold is folded in the high dimensional space. We call this method Tangent Distance Preserving Mapping (TDPM). TDPM uses tangent distance instead of geodesic distance, and then applies MDS to the tangent distance matrix to map the manifold into a low dimensional space in which we can get its nonlinear structure.

## 1. Introduction

In many practical applications, such as data mining, machine learning, computer vision, the dimensionality reduction is a necessary pre-processing step for the purpose of noise reduction and reducing the computation complexity. The basic principle of dimensionality reduction is to map the data point from the original space to a low dimensional space through a linear or a nonlinear map. Many dimensionality reduction methods have been developed to deal with this problem. Principal Component Analysis (PCA) [1] and Linear Discriminant Analysis (LDA) [2] are two traditional linear methods. PCA attempts to preserve the global structure of the data points' distribution, and LDA attempts to make the samples as separable as possible in the low dimensional space. But in the real world, lots of data are usually nonlinear and have high dimensionality, so using the traditional linear methods are difficult to find the true structure of the data points. Recently, some nonlinear algorithms have been developed, such as Locally Linear Embedding (LLE) ([3], [4]), ISOMAP [5], Local Tangent Space Alignment [6], etc. All these methods attempt to unfold the nonlinear manifold from the high dimensional space to the low dimensional embedding space. For instance, we apply ISOMAP to the SWISSROLL, the result is a 2-dimension plane which is shown in Fig.1 from which we can see that the manifold is unfolded to its essential dimension. Other algorithms have the similar results.

We must point out here that all these algorithms just unfold the manifold and do not reveal how the manifold is folded in the high dimensional space.

Our method is motivated by the upper problem. The start point of our method is to find a reasonable measure that can depict the relationship between each pair of the data points properly. We use tangent distance here, because tangent distance can reveal the nonlinear structure of the manifold trustily. Then we use MDS [7] to try to preserve this nonlinear structure, and map the data points to the low dimensional space.

The rest of the paper is organized as follows: In section 2, we explain how to compute the tangent distance between data points of a manifold in detail. In section 3, the whole algorithm is presented. Then in section 4, several experiments are conducted to demonstrate the effectiveness and robustness of our proposed method. Finally, in section 5 the conclusion is given.

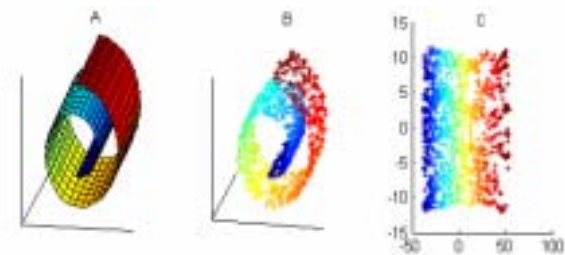

**Fig.1** 3-dimension data (B) is sampled from a 2-dimension manifold (A), and (C) is the result of applying ISOMAP to these data points.

## 2. Tangent Distance

Unlike Euclidean distance which computes the linear distance between two points, the tangent distance ([8], [9]) computes the minimum distance between a data point and the other data point's tangent space. In this section we'll describe how to compute the tangent distance between two data points on the manifold.

### 2.1 How to Compute the Tangent Space

Usually the data points are discrete, so how to construct the approximate tangent space is a challenging problem. However, we borrow the idea of a famous nonlinear dimensionality reduction algorithm named Local Tangent Space Alignment (LTSA) [6] here. LTSA uses local PCA to get the tangent space of each data point.

Let F be a manifold embedded in $R^m$. We assume that the manifold is generated from a smooth enough generating function $f(\mu), \mu \in R^d$ (d<m).

The data points $x_i = f(\mu_i) \quad i = 1,...,n$ are consist of n m-dimensional vectors. Let $x_{i_j} \quad j = 1,...,k$ be the k nearest neighbours of $x_i$, and they form a matrix:

$X_i = [x_{i_1},...,x_{i_k}], x_{i_j} \in R^m (j = 1,...,k)$.

Since $f(\mu), \mu \in R^d$ is smooth enough, by first-order Taylor expansion at $\mu^*$, we have

$$f(\mu) = f(\mu^*) + J_f(\mu^*) * (\mu - \mu^*) + O(\|\mu - \mu^*\|^2) \quad (1)$$

Where $J_f(\mu^*) \in R^{m*d}$ is the Jacobi matrix of f at $\mu^*$. We can see that the tangent space is spanned by the columns of $J_f(\mu^*)$. So, we need to find a matrix whose columns are the orthogonal basis of the local tangent space instead of $J_f(\mu^*)$, because we can not get $J_f(\mu^*)$.

Since the tangent space we need is local, therefore we can use only the k nearest neighbors to construct it.
Notice that from (1) we have

$$f(\mu) \approx f(\mu^*) + J_f(\mu^*) * (\mu - \mu^*) \quad (2)$$

So, we can replace the symbols as follows:

$$x_{i_j} \approx x_i + P_i \alpha_{ij} \quad (3)$$

Here, $P_i$ is a matrix comprised of the orthogonal basis of the local tangent space. $\alpha_{ij}$ is the coordinate of $x_i$'s j-th nearest neighbour in the local tangent space. Equation (3) means that we use the discrete data points to approximate the original smooth generating function $f(\mu)$. What we need to do is to make the error of (3) small enough. This is equal to solve the optimum problem:

$$\min_{x,P_i,\Phi} \sum_{j=1}^{k} \|x_{i_j} - (x + P_i \alpha_j)\|_2^2 \quad (4)$$

Where $\Phi = [\alpha_1, \alpha_2,...,\alpha_k]$, and $P_i$ is an orthonormal matrix that spans the approximate local tangent space of $x_i$. This optimum problem equates to PCA that is performed among the k nearest neighbours of $x_i$, and the optimal $P_i$ is given by the d left singular vectors of $X_i(I - ee^T/k)$ corresponding to its d largest singular values. This means that we use the local tangent space at the mean of $x_i$'s k nearest neighbours to approximate the local tangent space at $x_i$.

### 2.2 How to compute the tangent distance

In this paper, we use a distance metric that scales the minimum distance between a data point and the other point's tangent space, namely the tangent distance.
In section 2.1, we have showed how to get the tangent space of a given data point approximately. So, the tangent distance can be computed as follows:
Let $x^*$ be a given data point, and we assume the local tangent space at $x^*$ is spanned by $P^*$'s column vectors, namely $P^*$'s columns are the orthogonal basis of the local tangent space at $x^*$. For any given data point x, the distance between it and any point on $x^*$'s tangent space can be written as $\|(x^* + P^*\alpha) - x\|$, where $\alpha$ is a vector whose elements are the coordinates of the point on $x^*$'s tangent space. So, the tangent distance is the solution of the following optimum problem:

$$\min_{\alpha} \|(x^* + P^*\alpha) - x\| \quad (5)$$

This optimum problem is easy to solve, because it is a quadratic problem of $\alpha$.

## 3. Tangent Distance Preserving Mapping (TDPM)

Consider a data set represented by the columns of a matrix $X = [x_1, x_2,...,x_n] \in R^{m*n}$, $x_1, x_2,...,x_n$ are column vectors corresponding to the data points. The first step of our algorithm is to compute the tangent distance between each pair of the data points. Then in the interest of preserving the nonlinear structure of the manifold, we apply MDS to the tangent distance matrix to approximate the tangent distance by the Euclidean distance in the low dimensional embedding space. The concrete algorithmic procedure is formally stated as follows:
TDPM algorithm procedure:

**Input:** dataset $X \in R^{m*n}$, and the dimension of the mapping space: d.
**Output:** mapping vectors $Y \in R^{d*n}$.
**step1.** Find the K-nearest neighbours of each data point;
**step2.** Compute the tangent distance between each pair of points:

For each data point $x_i$:

a. Compute each point's local gram matrix $X_i$

b. Compute the eigenvalues and the eigenvectors of the matrix $X_i(I - ee^T/k)$,

c. Let matrix $P_i$ be the d left singular vectors of $X_i(I - ee^T/k)$ corresponding to its d largest singular values

d. Compute the tangent distance between $x_i$ and every data point of the data set by solving the optimum problem:

$$\min_{\alpha} \|(x_i + P_i\alpha) - x_j\|, j = 1,...,n$$

And store the results in the matrix TD.
**step3.** Apply MDS to matrix TD to get the mapping points.

## 4. Experiments and Discussions

In this section, we evaluate our algorithm by applying it to two synthetic manifolds and a face manifold; and we also test our method's sensitivity to the parameter k and the number of the data points.

### 4.1 Synthetic manifolds

In this experiments, we choose k=12. Fig.2 and Fig.3 are the results of applying TDPM to Swiss roll and S-Curve. The number of the data point is 500 and 1000 respectively.

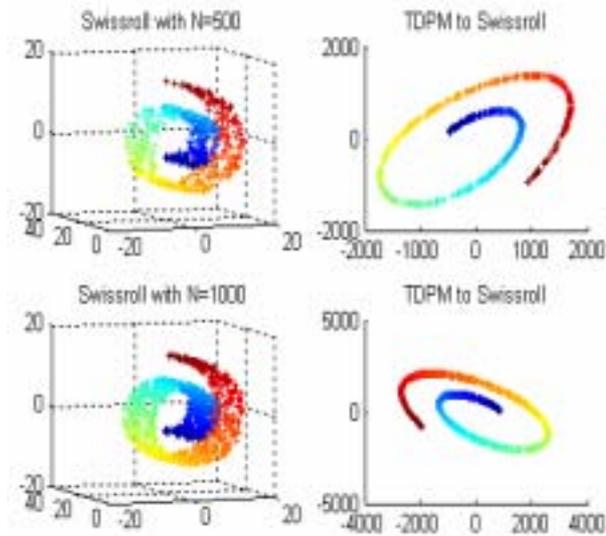

**Fig.2.** the mapping results of applying TDPM to SWISSROLL. The number of the data points is 1000 (upper) and 2000 (below).

The left picture is the data points generated from the original manifold, and the right picture is the mapping result of TDPM.

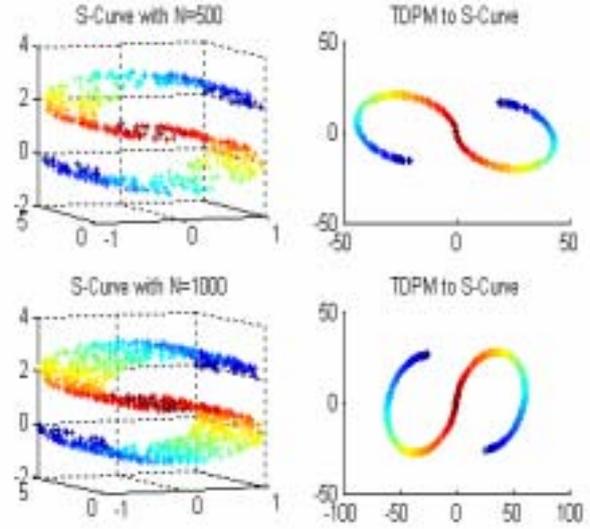

**Fig.3.** the mapping results of applying TDPM to SCURVE. The number of the data points is 1000 (upper) and 2000 (below). The left picture is the data points generated from the original manifold, and the right picture is the mapping result of TDPM.

From the result, we can see that TDPM can preserve the global geometric structure to a certain degree. The mapping results show how the manifolds are folded in the high dimensional space.

Meanwhile, we must notice that our method's results are somewhat scaled compared to the original manifolds. This is because we use the Euclidian distance in the embedding space to approximate the original tangent distance. Assume that point A and point B are any two data points. The tangent distance is less than or equal to the Euclidian distance between A and B. Only when the beeline connecting A and B is vertical to the tangent space of A or B, the tangent distance is equal to the Euclidian distance. So, when we map the data points to the low dimensional space, the tangent distance between each pair of data point may become shorter than it is in the original high dimensional space, and therefore the data points may distribute more spread than the original manifold.

### 4.2 Test the sensitivity to the parameter k and the number of the data points

In this experiment we test our method's sensitivity to the parameter k, namely the numbers of the nearest neighbours we choose when computing the tangent space, and the number of the data points. The results are shown below (Fig.4 and Fig.5):

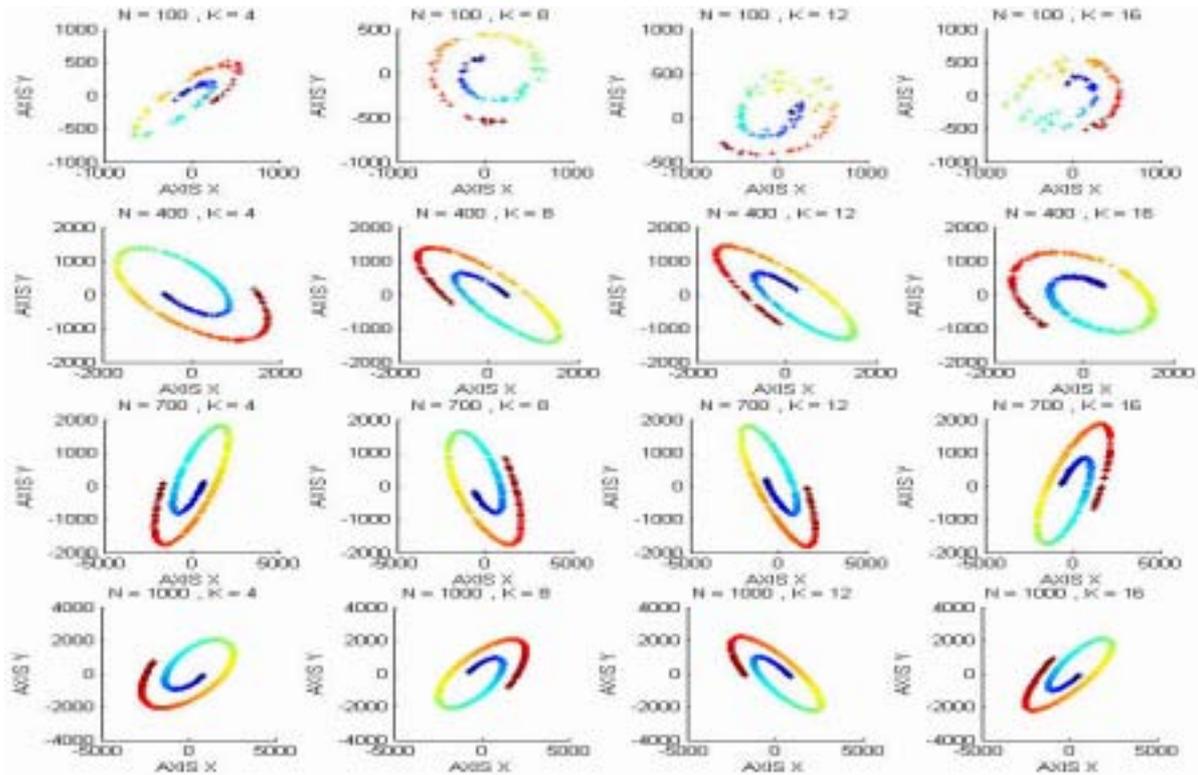

Fig.4 Using TDPM with different k to map the Swiss Roll to the two-dimensional space. The numbers of the data points are 100, 400, 700 and 1000 from top to bottom.

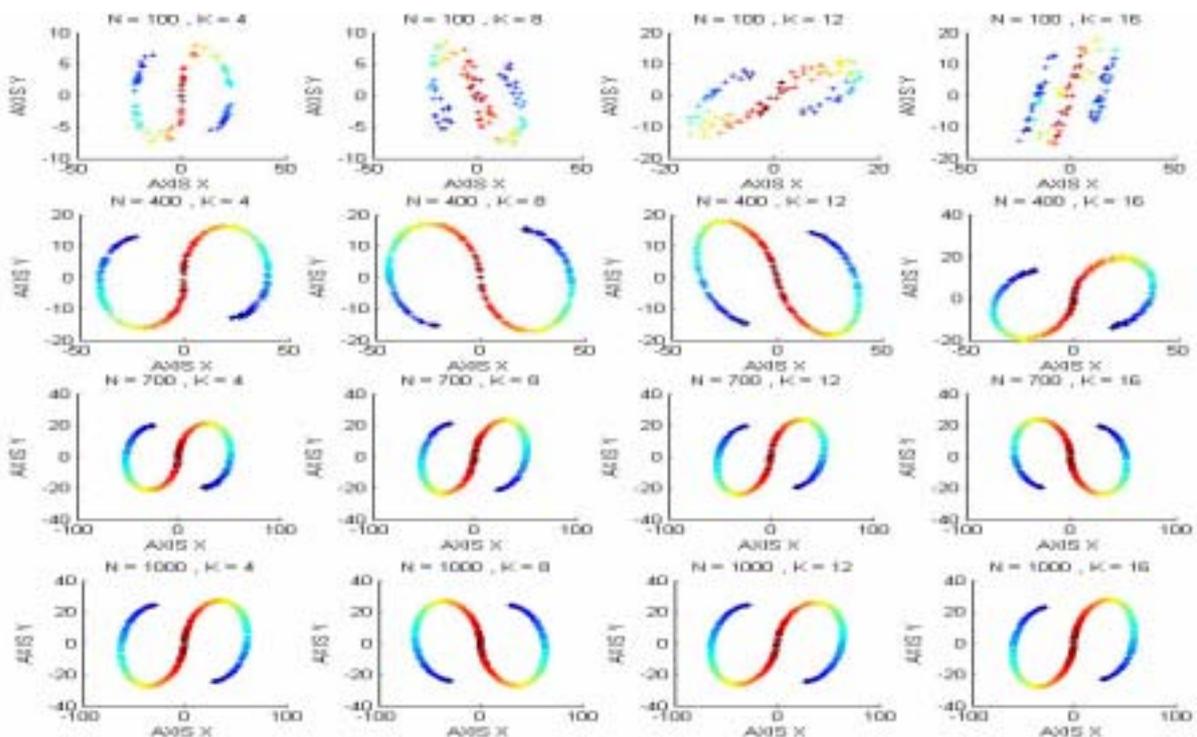

Fig.5 Using TDPM with different k to map the S-Curve to the two-dimensional space. The numbers of the data points are 100, 400, 700 and 1000 from top to bottom

From the results we can see that along with the decrease of the value k, the mapping results of TDPM does not change a lot. This means that our method is not sensitive to the number of the nearest neighbours to some extend, namely, the construct of the local tangent space is not sensitive to the numbers of the neighbours chosen at each data point.

Also notice that if the number of the data points is small, the mapping results may not converge to the true nonlinear structure of the manifold well, which can be explained as follows:

Since the data points are sparse, so when we compute the loacal tangent space, the results may distort, which results in the inaccuracy of the tangent distance. However, when the data points are enough at some degree, the mapping results do not change acutely.

**4.3 Face manifold**

Here we use a famous face images dataset [10] which contains 698 virtual faces with different degrees of freedom including rotation/pose and lighting/shading. Because our algorithm is to preserve the nonlinear structure of the manifolds, we first use TDPM to the face manifold to get a h-dimensionality embedding space, and we assume that the nonlinear structure of the manifold is embedding into this h-dimensionality space, then ISOMAP is applied to validate if TDPM has successfully achieved the goal, which means that if ISOMAP can successfully unfold the face manifod in the 2-dimension subspace, we can assure that TDPM indeed preserves the nonlinear structure of the original face manifold in the h-dimension subspace. In our experiment, we choose h equals 6 experimentally.
Fig.6 illustrates the results:

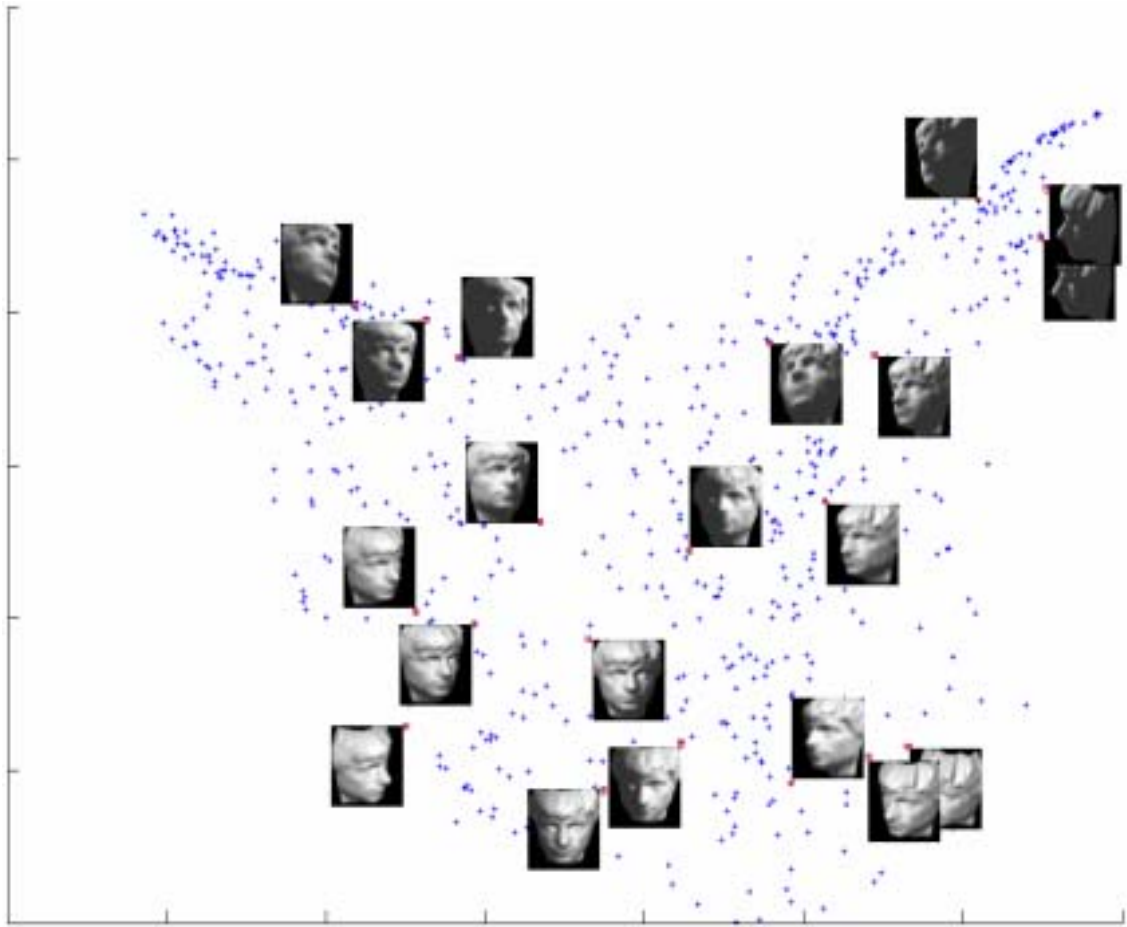

**Fig.6.** First using TDPM to map face manifold to a 6-dimensionality subspace, and then applying ISOMAP to map the result of TDPM to a 2-dimensionality subspace. Every red square corresponds to a face image.

From the result, we can see that the images are aligned in the embedding space according to the intrinsic degrees of freedom. The x-axis indicates the left-right rotation of the head, and the y-axis indicates the up-down rotation of the head, which means that ISOMAP successfully unfold the face manifold and also TDPM certainly preserves the nonlinear structure of the face manifold in some degree.

**5. Conclusions and future works**

In this paper a nonlinear structure preserving algorithm named Tangent Distance Preserving Mapping (TDPM) is proposed. Unlike usual manifold learning algorithms whic-h unfold the manifold in the low dimensional space, TDPM finds the nonlinear structure of the manifolds, and reveals how the manifolds are folded in the high dimensional space. Several experiments have been conducted to demonstrate the effectiveness and the robustness of our proposed method.

However, there are some farther works need to do, such as the scale problem and how to run TDPM together with some linear techniques. Also we will combine TDPM with the label information to design some categorization systems.